\documentclass[sigconf, screen, nonacm]{acmart}

\usepackage{xcolor}

\usepackage{multirow}
\usepackage{algorithm}
\usepackage{colortbl}
\usepackage{wrapfig}
\usepackage{hyperref}
\usepackage{cleveref}
\usepackage{pifont}
\usepackage{algpseudocode}
\usepackage{booktabs}
\usepackage{multirow}

\AtBeginDocument{%
  }

\setcopyright{acmlicensed}
\copyrightyear{2018}
\acmYear{2018}
\acmDOI{XXXXXXX.XXXXXXX}
\acmConference[Conference acronym 'XX]{Make sure to enter the correct
  conference title from your rights confirmation email}{June 03--05,
  2018}{Woodstock, NY}
\acmISBN{978-1-4503-XXXX-X/2018/06}

\begin{document}

\title{
A Utility-preserving De-identification Pipeline for Cross-hospital Radiology Data Sharing
}










\author{
Chenhao Liu$^1$, Zelin Wen$^1$, Yan Tong$^1$, Xinyu Tian$^2$, Yuchi Liu$^2$, Junjie Zhu$^3$, Ashu Gupta$^4$, Syed Mohammed Shamsul Islam$^5$, Tom Gedeon$^6$, Yue Yao$^{\dagger 1}$
}

\affiliation{
\vspace{6pt}
\institution{$^1$Shandong University, 
$^2$Australian National University, 
$^3$Hong Kong University of Science and Technology, 
$^4$Curtin University, 
$^5$Edith Cowan University, 
$^6$University of Western Australia}
 \country{}
}

\renewcommand{\shortauthors}{Trovato et al.}



\begin{abstract}

Large-scale radiology data are critical for developing robust medical AI systems. However, sharing such data across hospitals remains heavily constrained by privacy concerns. Existing de-identification research in radiology mainly focus on removing identifiable information to enable compliant data release. Yet whether de-identified radiology data can still preserve sufficient utility for large-scale vision-language model training and cross-hospital transfer remains underexplored. In this paper, we introduce a utility-preserving de-identification pipeline (UPDP) for cross-hospital radiology data sharing. Specifically, we compile a blacklist of privacy-sensitive terms and a whitelist of pathology-related terms. For radiology images, we use a generative filtering mechanism that synthesis a privacy-filtered and pathology-reserved counterparts of the original images. These synthetic image counterparts, together with ID-filtered reports, can then be securely shared across hospitals for downstream model development and evaluation. Experiments on public chest X-ray benchmarks demonstrate that our method effectively removes privacy-sensitive information while preserving diagnostically relevant pathology cues. Models trained on the de-identified data maintain competitive diagnostic accuracy compared with those trained on the original data, while exhibiting a marked decline in identity-related accuracy, confirming effective privacy protection. In the cross-hospital setting, we further show that de-identified data can be combined with local data to yield better performance.

\end{abstract}


\maketitle

\section{Introduction}

\begin{figure}
\centering
\includegraphics[width=1\linewidth]{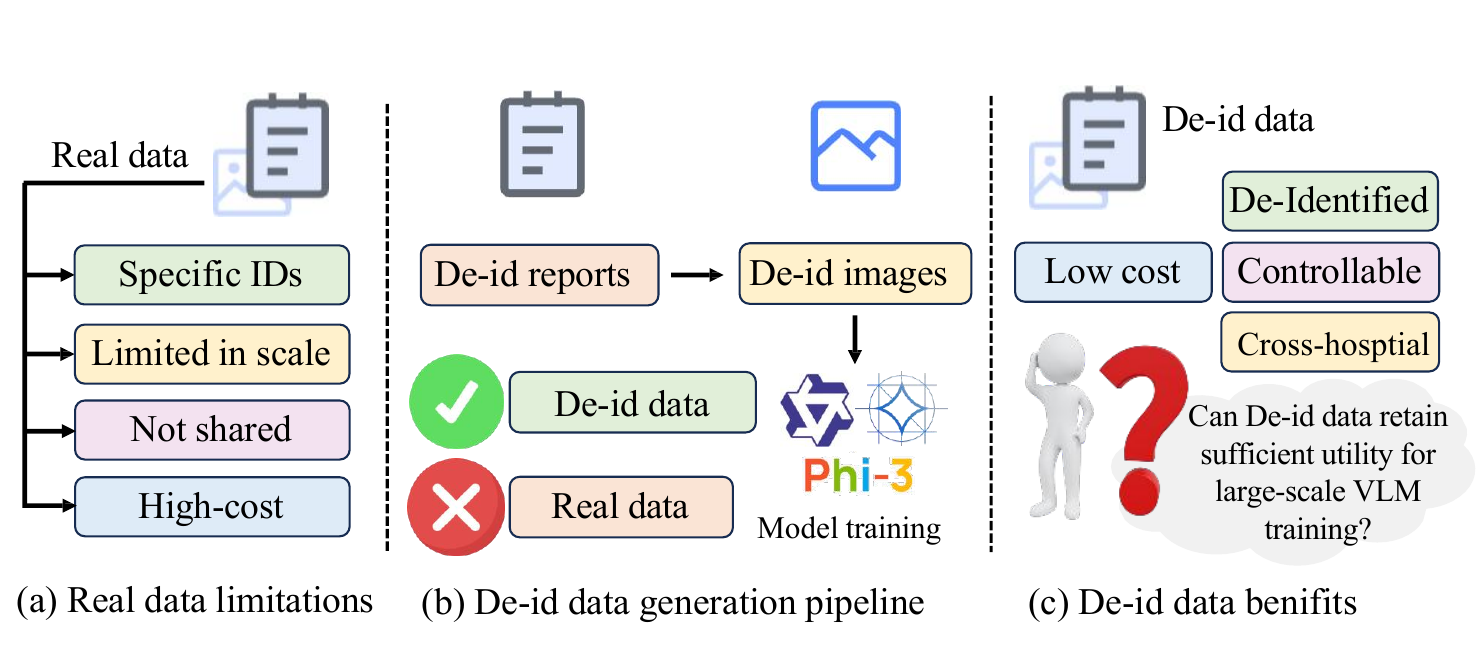}
\vspace{-8mm}
\caption{
Limitations of real radiology data for cross-hospital model training. Real radiology data are usually difficult to share across hospitals due to privacy restrictions, limited accessibility, and institution-specific barriers. These challenges motivate de-identification as a practical solution for privacy-preserving data transfer and scalable cross-hospital model training.
}
\vspace{-1.5em}
\label{fig:introd}
\end{figure}


Radiology data play a crucial role in the development of medical Vision Language Models (VLMs), facilitating applications such as automated report generation, diagnostic assistance, and large-scale clinical studies \cite{tanno2025collaboration, el2011systematic}. However, sharing data across hospitals is usually hindered by strict privacy rules~\cite{kaissis2020secure,geis2021datasharing}. Additionally, the training of VLMs for generating high-quality reports requires substantial human and computational resources \cite{irvin2019chexpert}. These challenges limit the size and diversity of public available datasets and impede cross-hospital collaboration, thereby slowing progress in research and model development.


To address this issue, prior de-identification studies in radiology have mainly focused on removing explicit identifiers from text or masking privacy-sensitive information for compliant data release~\cite{dutt2025devilpromptsdeidentificationtraces,neamatullah2008automated}. While such efforts are important, they largely emphasize privacy removal itself. For cross-hospital medical AI development, however, privacy protection alone is not sufficient. The shared data should also preserve enough clinically relevant information to remain useful for downstream training. This issue becomes even more important in the era of large-scale VLM training, where both image content and report semantics contribute to representation learning. If de-identification removes not only identity-related information but also diagnostically important pathology cues, the resulting data may be safe to share but substantially less useful for model development. Therefore, a key open question is whether radiology data can be de-identified in a way that simultaneously protects privacy and preserves utility for large-scale model training and cross-hospital transfer.

In this paper, we study this problem from a utility-preserving perspective and propose a Utility-Preserving De-identification Pipeline (UPDP) for cross-hospital radiology data sharing. Our main idea is to explicitly filter out privacy-sensitive information from clinically relevant pathological data, and then preserve the latter during the de-identification process. On the senmantic side, we construct a blacklist of privacy-sensitive terms as well as a whitelist of pathology-related terms to filter reports while retaining clinically meaningful semantics. On the image side, instead of directly discarding or heavily masking original data, we introduce a generative filtering scheme to produce privacy-filtered yet pathology-preserved image counterparts. These de-identified image-report pairs can then serve as safer transferable data for downstream model development across institutions.

We evaluate our method on publicly available chest X-ray datasets MIMIC-CXR-JPG \cite{johnson2019mimic} and IU-Xray \cite{DemnerFushman2016RadiologyCollection} by training vision-language models to generate radiology reports. Results show that models trained on de-identified data produce reports with diagnostic accuracy comparable to those trained on the original data, while significantly reducing identity-related information. This confirms the effectiveness of our privacy protection approach. In cross-hospital settings, combining de-identified data with local datasets further improves model report generation quality and downstream diagnostic performance, demonstrating the practical utility and generalization of our method.

\section{Related Work}

\textbf{De-identification in medical image.} De-identification aims to remove patient-identifiable information from medical data, including both textual metadata and image content, to support safe data sharing.~\cite{moore2015identification} In the textual domain, conventional approaches typically rely on rule-based systems, curated dictionaries, and named entity recognition models to detect protected health information; however, their applicability is often limited due to variations in reporting styles and terminology. In the visual domain, traditional de-identification techniques typically rely on metadata sanitization, anatomical masking, or spatial obfuscation (\textit{e.g.}, blurring, inpainting) to obscure identity-related cues; while straightforward, these heuristic operations often degrade radiographic fidelity and inadvertently suppress diagnostically critical features. Recent advances have shifted toward generative frameworks, such as GANs and diffusion models, to synthesize privacy-compliant images with enhanced perceptual realism; however, they often prioritize privacy at the cost of diagnostic fidelity and cross-modal semantic alignment between images and their paired report, which limits their effectiveness for downstream vision-language tasks.

\textbf{Medical Report Generation.}  Medical report generation focuses on automating the creation of clinically coherent radiology reports from imaging modalities such as chest X-rays. Early approaches predominantly employed CNN-RNN architectures~\cite{shin2016learning,yuan2019automatic}, wherein convolutional encoders extracted visual representations and recurrent decoders generated text autoregressively~\cite{monshi2020deep, pang2023survey}. Subsequent methods incorporated attention mechanisms, hierarchical decoders, and memory modules to better model long-range dependencies and enhance the structure of reports~\cite{you2021aligntransformer}. More recently, transformer-based models, such as R2Gen and HuatuoGPT-Vision, have demonstrated superior capabilities in modeling global context and producing fluent reports by leveraging large-scale pre-training ~\cite{yang2022knowledge,chen2024huatuogpt,chen2020generating}. Furthermore, strategic prompt engineering has been shown to enhance the logical consistency and clinical accuracy of generated outputs~\cite{yao2025simple}. Despite these advances, all methods remain critically dependent on large-scale, high-quality image-report pairs, and their performance tends to degrade when training data is scarce. This limitation is primarily due to the relative scarcity of medical imaging data, as well as the specific credentials required to access such data~\cite{mamdouh2025advancements}.

\begin{figure*}[t]
    \centering
    \captionsetup{skip=1pt}
    \includegraphics[width=\textwidth]{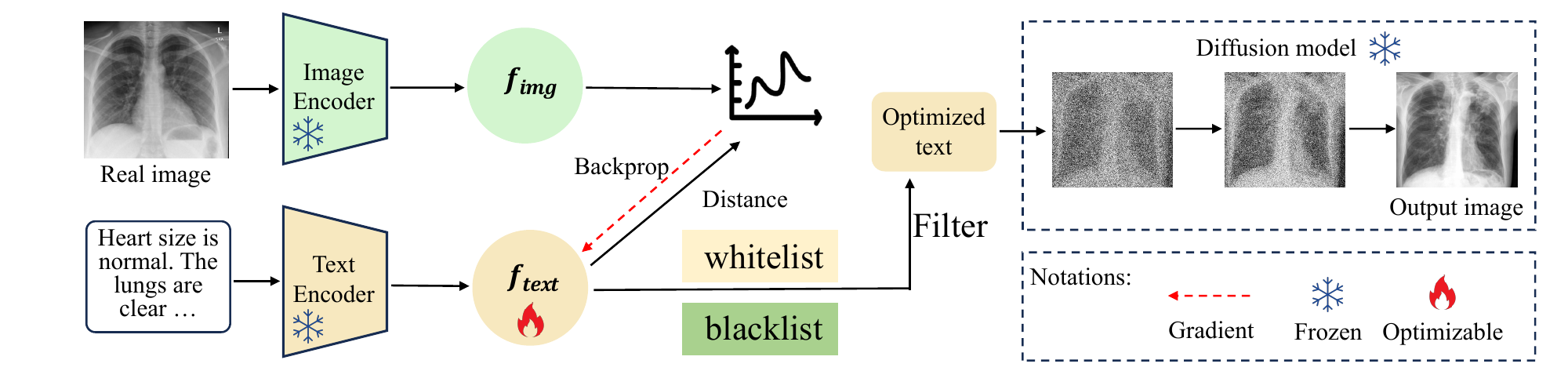}
    \caption{
    Overview of the utility-preserving De-identification pipeline. Given an real image-report pair, we extract image and text features with a pretrained vision language model. We optimize continuous content embeddings toward the paired image features, as well as a semantic optimization towards  whitelist and blacklist, and use the resulting prompt to condition the diffusion model for chest X-ray de-identification.
    }
    \label{fig:pipeline}
    \vspace{-1em}
\end{figure*}

\textbf{Synthetic Data for Report Generation.} Given the scarcity of paired CXR-report data, researchers have explored using synthetic image-text pairs to augment training~\cite{liu2024can, yao2023attribute}. A common method is to generate synthetic CXRs from real reports using generative adversarial networks (GANs) or latent diffusion models (LDMs)~\cite{bluethgen2025vision,moris2024adapted}. These synthetic data can then be used to pre-train or fine-tune report generation frameworks, reducing dependence on real medical images and easing privacy concerns. While leveraging synthetic data has proven effective in improving model robustness~\cite{khosravi2024synthetically,ktena2024generative}, ensuring strict image-text semantic correspondence remains a fundamental bottleneck~\cite{chen2024medical}. Conditioning generation on noisy or overly sanitized clinical text—as frequently occurs after standard de-identification—can lead to the omission of critical pathological cues. Consequently, the resulting synthetic images may lack diagnostic fidelity, undermining their utility for training reliable report generation systems~\cite{chen2024medical, koetzier2024generating, hosseini2025synthetic}.

\section{Method}
\subsection{Motivation}
Radiology report generation relies on large-scale paired image-report datasets, yet privacy regulations severely constrain cross-institutional data sharing. Medical images usually contain plenty of identifiable information, either explicitly through burned-in annotations or implicitly via anatomical features on radiology images. Furthermore, reports may retain residual identifiers even after standard de-identification procedures~\cite{Schwarz2019FaceRecMRI,Carrell2013HIPS}. Regulations such as HIPAA and GDPR impose strict rules on data use and dissemination~\cite{McGraw2013HIPAA,GDPR}, which significantly restrict access to high-quality training data. Existing methods largely focus on removing identifiable information, often neglecting clinically relevant features that are important for downstream tasks and model performance. In radiology report generation, this creates a challenge: how to remove privacy-sensitive content while maintaining accurate image-text alignment and preserving diagnostic information across modalities. 

To address this, we treat de-identification as a controlled generation problem. Instead of modifying images or reports in isolation, we use diffusion-based image synthesis guided by de-identified textual prompts to produce coherent outputs. Specifically, for a target chest X-ray dataset, our objective is to generate a de-identified version that suppresses identity-related information while retaining clinically meaningful content for report training. We optimize textual prompts before synthesis rather than relying on raw reports, which allows precise and interpretable control over the balance between privacy protection and data utility.

\begin{algorithm}[t]    
\caption{De-identified Images Generation with Semantic Constrained list}     
\label{alg:DeIDPromptTuning}      
\noindent\rule{\linewidth}{0.5pt}    
\begin{algorithmic}[1]     
\Require          
$\mathcal{D}_T = \{(x_i, y_i)\}_{i=1}^{m_t}$,          
pretrained encoders $E_T, E_I$,          
diffusion model $G$,          
vocabulary $\mathcal{V}$,          
forbidden tokens $\mathcal{F}$ (blacklist),         
preferred tokens $\mathcal{P}$ (whitelist),          
learning rate $\eta$,          
top-K,          
num steps $T$     
\Ensure $\mathcal{D}_S$ (de-identified images dataset)      

\Statex \noindent\rule{\linewidth}{0.5pt}      

\State Initialize $\mathcal{D}_S \leftarrow \emptyset$      

\For{each $(x_i, y_i) \in \mathcal{D}_T$}         
    \State Tokenize report $y_i$: $H_i \leftarrow [e(w_1),..., e(w_L)]$         
    \State Extract initial semantic features:         
    \Statex \quad $f_{rpt} \leftarrow E_T(H_i)$, \quad $f_{img} \leftarrow E_I(x_i)$          

    \For{$t = 1$ to $T$}              
        \State \textbf{Semantic Alignment:} optimize content to preserve diagnostic content             
        \State Compute alignment loss: $\mathcal{L} \leftarrow 1 - \frac{f_{rpt} \cdot f_{img}}{\|f_{rpt}\| \|f_{img}\|}$             
        \State Update soft prompt embeddings: $f_{rpt} \leftarrow f_{rpt} - \eta \nabla_{f_{rpt}} \mathcal{L}$         
    \EndFor          

    \State \textbf{Semantic Constrained Projection (De-Identification)}
    \For{$j = 1$ to $L$}             
        \State Compute similarity with vocab: $s_j(t) = \cos(f_{rpt,j}, e(t)), \forall t \in \mathcal{V}$             
        \State \textbf{Blacklist filtering:} $s_j(t) = -\infty$ for $t \in \mathcal{F}$  \Comment{remove private info}             
        \State \textbf{Whitelist top-K selection:} $\mathcal{T}_j \leftarrow \text{TopK}(\{s_j(t) \mid t \in \mathcal{P}\})$ \Comment{keep diagnostic info}             
        \State Sample discrete token: $\tilde{y}_{i,j} \sim \text{softmax}(s_j(t))$ for $t \in \mathcal{T}_j$         
    \EndFor          

    \State Construct de-identified content: $\tilde{y}_i \leftarrow [\tilde{y}_{i,1}, ..., \tilde{y}_{i,L}]$         
    \State Generate de-identified image: $\tilde{x}_i \leftarrow G(\tilde{y}_i)$         
    \State Add to de-identified images dataset: $\mathcal{D}_S \leftarrow \mathcal{D}_S \cup \{(\tilde{x}_i, y_i)\}$     
\EndFor      

\State \Return $\mathcal{D}_S$   
\end{algorithmic}    
\noindent\rule{\linewidth}{0.5pt}   
\vspace{-2.0em} 
\end{algorithm}

\subsection{Problem Definition} 

We formulate medical image de-identification as a controlled image generation task. Rather than disseminating raw data, our objective is to transform a target chest X-ray dataset into a de-identified counterpart that systematically suppresses privacy-sensitive cues while preserving diagnostically relevant clinical content.

Formally, let
\begin{equation}
\mathcal{D}_T = \{(x_i, y_i)\}_{i=1}^{m_t}
\end{equation}
denote the target dataset, where $x_i$ is a CXR image and $y_i$ is the corresponding radiology report. The reports are drawn from publicly available, HIPAA-compliant corpora (\textit{e.g.}, MIMIC-CXR and IU X-Ray), but may still contain residual identifiers or dataset-specific biases.

The objective is to construct a de-identified dataset
\begin{equation}
\mathcal{D}_S = \{(\tilde{x}_i, y_i)\}_{i=1}^{m_s},
\end{equation}
where $\tilde{x}_i$ denotes a privacy-preserving version of $x_i$. Specifically, $\tilde{x}_i$ is generated by a latent diffusion model conditioned on an optimized content $\tilde{y}_i$, such that identity-related features are suppressed while maintaining semantic alignment with $y_i$.

To achieve this, we adopt a lexically constrained content optimization framework with three stages. First, each image report pair is encoded using a pretrained vision language model to obtain aligned multimodal representations. Next, the report embeddings are refined to better match the paired image in feature space, while lexical constraints are applied to suppress sensitive tokens and emphasize clinically informative terms. Finally, the optimized conteng is used to guide a diffusion model to generate a de-identified image, which is then paired with the original report for downstream report generation.

\textbf{De-identified Image Generation.}  
The resulting content $\tilde{y}_i$ is provided to a latent diffusion model $G$ to generate the image:
\begin{equation}
\tilde{x}_i = G(x_i, \tilde{y}_i).
\end{equation}
This procedure enables controlled image synthesis guided by semantically aligned prompts while preserving diagnostic content.

\begin{figure*}
\centering
\setlength{\abovecaptionskip}{0.5em} 
\includegraphics[width=1\linewidth]{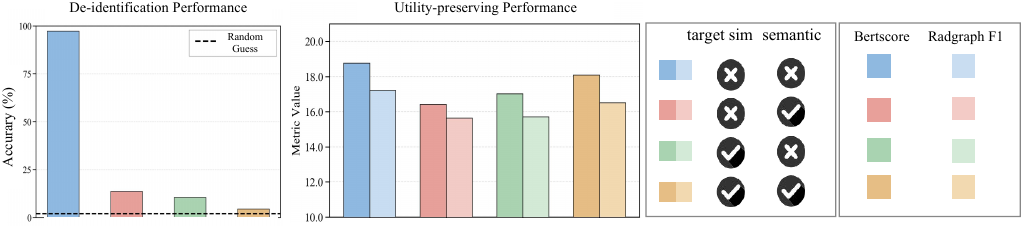 }
\vspace{-5mm}
\caption{(Left): Effectiveness of De-identification. (Right): Effectiveness of utility-preserving evaluation using Bert Score and RadGraph F1. We show that our UPDP pipeline can successfully filer our ID information while still preserving utility of training VLM model. }
\vspace{-1.5em}
\label{fig:introduct}
\end{figure*}
\subsection{Semantic Filtering for De-identification}

To suppress identity-related information while preserving diagnostic content, lexical constraints are applied in the mapping from continuous embeddings to discrete tokens.

\paragraph{Constrained Token Selection.}  
Token selection is performed in two stages: global exclusion using a blacklist and soft promotion using a whitelist.

All sensitive tokens in a blacklist $\mathcal{F}$ are excluded prior to the computation of Top-$K$ candidates: \begin{equation} s_{i,j}(t) = -\infty, \quad \forall t \in \mathcal{F}. \end{equation}The blacklist is designed to cover multiple categories of sensitive information, including direct patient identifiers (\textit{e.g.}, name, medical record number), healthcare personnel identifiers (\textit{e.g.}, doctor, attending), contact and location information, dates, demographic attributes, institutional identifiers, and other common PHI cues according to HIPAA guidelines. By enforcing this global hard constraint, identity-sensitive tokens are deterministically excluded from generation.

The Top-$K$ candidates $\mathcal{T}_{i,j}$ are then selected from the remaining tokens:
\begin{equation}
\mathcal{T}_{i,j} = \text{TopK}(s_{i,j}, K).
\end{equation}

Within these Top-$K$ candidates, tokens present in a whitelist $\mathcal{P}$ are given a small positive bias $\lambda$ to favor clinically informative and diagnostically relevant terms. The whitelist typically includes terms describing imaging modalities, acquisition views, core anatomical structures, tissue characteristics, and neutral or normal clinical descriptors. This soft promotion mechanism enhances the likelihood of selecting medically useful tokens while maintaining privacy:
\begin{equation}
\tilde{s}_{i,j}(t) = s_{i,j}(t) + \lambda \cdot \mathbf{1}_{t \in \mathcal{P}}, \quad t \in \mathcal{T}_{i,j}.
\end{equation}

Finally, discrete token indices are obtained by sampling with softmax over $\tilde{s}_{i,j}$ restricted to $\mathcal{T}_{i,j}$, providing a balance between lexical control and diversity. This two-stage design ensures strict global suppression of sensitive tokens while simultaneously promoting clinically relevant tokens within the Top-$K$ candidates, enabling privacy-preserving yet diagnostically faithful generation.

\subsection{De-identification Optimization}

To preserve diagnostic utility during de-identification, textual embeddings are optimized while maintaining semantic consistency with the original image content.

\paragraph{Semantic Feature Alignment.}  
Given an image-report pair $(x_i, y_i)$, textual embeddings $\tilde{H}_i \in \mathbb{R}^{L \times D}$ are optimized to align with the image representation in the vision-language latent space:
\begin{equation}
\mathcal{L}_i = 1 - \cos(E_T(\tilde{H}_i), E_I(x_i)).
\end{equation}
where $E_T$ and $E_I$ denote the text and image encoders of a pretrained vision-language model (\textit{e.g.}, BiomedCLIP). This objective encourages each token-level embedding to capture clinically relevant features from the image, preserving diagnostic information and ensuring semantic consistency with the associated report.

\paragraph{Nearest-Neighbor Projection.}  
To bridge continuous optimization with discrete generation, the optimized embeddings are projected onto the token embedding space using normalized nearest-neighbor matching. Each embedding vector $\tilde{e}_{i,j}$ is $\ell_2$-normalized and compared with the normalized token embedding matrix $E \in \mathbb{R}^{|\mathcal{V}| \times D}$:
\begin{equation}
s_{i,j}(t) = \cos(\tilde{e}_{i,j}, e(t)), \quad t \in \mathcal{V}.
\end{equation}
where $e(t)$ denotes the embedding of token $t$.

After computing scores for all tokens, forbidden tokens in a blacklist $\mathcal{F}$ are excluded by assigning them a score of $-\infty$. The Top-$K$ candidates are then selected based on the remaining scores:
\begin{equation}
\mathcal{T}_{i,j} = \text{TopK}(s_{i,j}, K).
\end{equation}
Within these Top-$K$ candidates, tokens in a whitelist $\mathcal{P}$ are given a small positive bias to favor clinically informative terms. A final token is chosen from $\mathcal{T}_{i,j}$ using either softmax sampling or greedy selection, producing discrete indices that are mapped back to the embedding space for generation.

\section{Results}

\begin{table*}[t]
\centering
\footnotesize
\caption{Utility-preserving performance on MIMIC and IU X-Ray. We evaluate two distinct application scenarios: (1) training VLMs exclusively on transferred data, and (2) using transferred data to augment local datasets. In both settings, we systematically compare models trained on de-identified versus non-de-identified data to assess the pipeline's utility-preserving capability.}
\vspace{-3mm}


\label{tab:main_results1}

\begin{tabular}{c | c | c | c | c c c c c c c c} 
\toprule
\multirow{2}{*}{\textbf{Dataset}} & \multicolumn{2}{c|}{\textbf{Transfer}} 
& \multirow{2}{*}{\textbf{Local}}
& \multicolumn{8}{c}{\textbf{Metrics}} \\
\cmidrule(lr){2-3} 
\cmidrule(lr){5-12}
& \textbf{Target Sim} & \textbf{Semantic} 
& & \textbf{BLEU-1} & \textbf{BLEU-2} & \textbf{BLEU-3} & \textbf{BLEU-4} 
& \textbf{METEOR} & \textbf{ROUGE-L} & \textbf{BERTScore} & \textbf{RadGraph F1} \\
\midrule

\multirow{8}{*}{\textbf{MIMIC-CXR~\cite{johnson2019mimic}}}
& \ding{55} & \ding{55} & \ding{51} & 18.42 & 10.09 & 5.84 & 3.68 & 8.88 & 17.32 & 18.76 & 17.22 \\
\cmidrule(lr){2-12}
& \ding{55} & \ding{55} & \ding{55} & 11.34 & 5.64 & 3.29 & 2.09 & 6.68 & 15.94 & 16.42 & 15.65 \\
& \ding{51} & \ding{55} & \ding{55} & 12.62 & 6.74 & 4.15 & 2.83 & 7.60 & 17.83 & 17.03 & 15.70 \\
& \ding{51} & \ding{51} & \ding{55} & \textbf{13.62} & \textbf{7.39} & \textbf{5.29} & \textbf{3.49} & \textbf{7.89} & \textbf{17.88} & \textbf{18.01} & \textbf{16.51} \\
\cmidrule(lr){2-12}

& \ding{55} & \ding{55} & \ding{51} & 19.61 & 10.52 & 6.05 & 3.24 & 9.12 & 17.38 & 17.24 & 16.21 \\
& \ding{51} & \ding{55} & \ding{51} & 20.01 & 11.03 & 5.92 & 3.37 & 9.17 & 17.68 & 18.75 & 17.42 \\
& \ding{51} & \ding{51} & \ding{51} & \textbf{23.93} & \textbf{11.79} & \textbf{6.45} & \textbf{3.80} & \textbf{9.35} & \textbf{18.29} & \textbf{20.65} & \textbf{18.65} \\
\midrule

\multirow{8}{*}{\textbf{IU X-Ray~\cite{demner2015preparing}}}
& \ding{55} & \ding{55} & \ding{51} & 45.47 & 30.05 & 20.74 & 13.98 & 21.11 & 33.84 & 41.01 & 41.48 \\
\cmidrule(lr){2-12}
& \ding{55} & \ding{55} & \ding{55} & 42.78 & 28.30 & 17.70 & 12.39 & 19.96 & 32.33 & 37.58 & 37.01 \\
& \ding{51} & \ding{55} & \ding{55} & 43.57 & 28.41 & 19.45 & 13.48 & 20.90 & 32.89 & 38.04 & 37.65 \\
& \ding{51} & \ding{51} & \ding{55} & \textbf{44.42} & \textbf{31.58} & \textbf{19.90} & \textbf{13.84} & \textbf{21.08} & \textbf{33.58} & \textbf{39.10} & \textbf{38.14} \\
\cmidrule(lr){2-12}

& \ding{55} & \ding{55} & \ding{51} & 45.37 & 29.89 & 20.97 & 14.58 & 21.38 & 34.27 & 39.02 & 39.69 \\
& \ding{51} & \ding{55} & \ding{51} & 46.95 & 31.29 & 21.27 & 15.03 & 22.15 & 35.25 & 40.67 & 41.91 \\
& \ding{51} & \ding{51} & \ding{51} & \textbf{47.01} & \textbf{31.58} & \textbf{22.27} & \textbf{15.83} & \textbf{22.45} & \textbf{34.50} & \textbf{41.28} & \textbf{42.57} \\
\bottomrule
\end{tabular}
\end{table*}

\subsection{Experiment Details}

Experiments are conducted on two widely used chest X-ray benchmarks: IU X-Ray~\cite{DemnerFushman2016RadiologyCollection} and MIMIC-CXR~\cite{johnson2019mimic}. 
The IU X-Ray dataset contains 8,121 chest radiographs paired with 3,996 diagnostic reports, whereas MIMIC-CXR comprises 377,110 radiographs from 227,835 imaging studies of 65,379 patients, each associated with a free-text radiology report. 
We follow the official train/validation/test splits provided by the original datasets to ensure fair comparison with prior work.

To maintain a consistent generation and evaluation protocol across all methods, we restrict both datasets to frontal-view (PA) radiographs only. 
This standardization mitigates cross-view distribution shifts and enforces anatomical consistency during both synthetic data generation and report evaluation, which is critical for preserving diagnostic utility. 
For MIMIC-CXR, we further sample a 10k subset from the training split to enable controlled and computationally efficient experimentation while preserving sufficient data diversity. 
This setup allows us to systematically evaluate whether de-identified data can retain sufficient utility for downstream model training under constrained data regimes.

Qwen2.5-VL~\cite{bai2025qwen2} serves as the backbone model for evaluations on both IU X-Ray and MIMIC-CXR. To assess the impact of model architecture and capacity, we further fine-tune Gemma-3-4B-IT~\cite{team2025gemma} and Phi-3.5-Vision-Instruct~\cite{abdin2024phi} on the IU X-Ray dataset. All models are adapted via parameter-efficient Low-Rank Adaptation (LoRA)~\cite{hu2022lora} for 3 epochs, with hyperparameters set as follows: $\text{lora\_rank}=64$, $\text{lora\_alpha}=64$, $\text{lora\_dropout}=0.05$, base learning rate $1\times10^{-5}$, merged-layer learning rate $1\times10^{-5}$, and vision backbone learning rate $1\times10^{-4}$. A uniform batch size of 36 is used across all experiments. This configuration balances stable convergence with computational efficiency on both datasets.

We further compare our method against representative content optimization baselines on the IU X-Ray dataset, including MMA-Diffusion~\cite{yang2024mma} and STEPS~\cite{qiu2025steps}. 
To ensure a fair and controlled evaluation, all methods are implemented and assessed under a unified setting using the same backbone model, \textit{Qwen2.5-VL-7B-Instruct}. 
Although these baselines were originally developed for general text-to-image optimization, we adapt them to the medical report-generation task, enabling a more meaningful comparison of their effectiveness in this domain.

Models are evaluated using standard medical report generation metrics, including BLEU-1 to BLEU-4~\cite{papineni2002bleu}, METEOR~\cite{banerjee2005meteor}, ROUGE-L~\cite{lin2004rouge}, 
as well as BERTScore~\cite{zhang2019bertscore} and RadGraph F1~\cite{jain2021radgraph}. 
These metrics collectively provide a comprehensive assessment of both lexical precision and semantic fidelity, 
as well as clinical correctness and entity-level alignment between generated reports and ground-truth references.

\vspace{-1.0em}

\subsection{Main Results}

\begin{table*}[t]
\centering
\vspace{-2mm}
\caption{ Utility-preserving performance on different models. We assume that a hospital wants to specialize a large model. We compare the performance of different vision-language models (VLMs) Finetuned on various datasets before and after specialization.}
\vspace{-3mm}
\footnotesize
\label{tab:main_results_iuxray_finetune}

\setlength{\tabcolsep}{1.2mm}
\begin{tabular}{ c | c | c | c | c | c c c c c c c c} 
\toprule
\multirow{2}{*}{\textbf{Model}} 
& \multirow{2}{*}{\textbf{Finetune}}
& \multicolumn{2}{c|}{\textbf{Transfer}} 
&\multirow{2}{*}{ \textbf{Local}}
& \multicolumn{8}{c}{\textbf{Metrics}} \\
 \cmidrule(lr){3-4}
 \cmidrule(lr){6-13}
& & \textbf{Target Sim} & \textbf{Semantic} 
& & \textbf{BLEU-1} & \textbf{BLEU-2} & \textbf{BLEU-3} & \textbf{BLEU-4} 
& \textbf{METEOR} & \textbf{ROUGE-L} & \textbf{BERTScore} & \textbf{RadGraph F1} \\
\midrule

\multirow{4}{*}{\textbf{Gemma-3-4B-IT}} 
& \ding{55} & \ding{55} & \ding{55} & \ding{55} & 18.87 & 8.52 & 4.29 & 2.15 & 14.35 & 15.74 & 10.51 & 13.95 \\
& \ding{51} & \ding{51} & \ding{55} & \ding{55} & 36.91 & 22.90 & 15.81 & 10.88 & 15.22 & 27.13 & 37.77 & 36.87 \\
& \ding{51} & \ding{51} & \ding{51} & \ding{55} & 39.42 & 24.89 & 17.09 & 11.89 & 17.01 & 33.01 & 38.74 & 38.75 \\
& \ding{51} & \ding{55} & \ding{55} & \ding{51} & \textbf{41.60} & \textbf{27.37} & \textbf{18.41} & \textbf{12.33} & \textbf{19.82} & \textbf{33.30} & \textbf{40.24} & \textbf{39.27} \\
\midrule

\multirow{4}{*}{\textbf{Phi-3.5-vision-instruct}} 
& \ding{55} & \ding{55} & \ding{55} & \ding{55} & 19.36 & 8.67 & 4.75 & 2.53 & 14.47 & 17.73 & 20.99 & 18.80 \\
& \ding{51} & \ding{51} & \ding{55} & \ding{55} & 39.03 & 25.19 & 16.94 & 10.59 & 17.82 & 32.15 & 38.38 & 40.33 \\
& \ding{51} & \ding{51} & \ding{51} & \ding{55} & 42.18 & 26.78 & 18.33 & 12.82 & 18.94 & 32.80 & 39.92 & 40.70 \\
& \ding{51} & \ding{55} & \ding{55} & \ding{51} & \textbf{43.06} & \textbf{26.98} & \textbf{19.16} & \textbf{13.10} & \textbf{19.24} & \textbf{34.65} & \textbf{40.87} & \textbf{41.42} \\
\midrule

\multirow{4}{*}{\textbf{Qwen2.5-VL-7B-Instruct}} 
& \ding{55} & \ding{55} & \ding{55} & \ding{55} & 19.95 & 9.89 & 5.43 & 2.76 & 17.02 & 19.16 & 20.10 & 22.08 \\
& \ding{51} & \ding{51} & \ding{55} & \ding{55} & 42.78 & 28.30 & 17.70 & 12.39 & 19.96 & 32.33 & 39.10 & 38.14 \\
& \ding{51} & \ding{51} & \ding{51} & \ding{55} & 44.42 & \textbf{31.29} & 19.90 & 13.84 & 21.08 & 33.58 & 39.01 & 40.48 \\
& \ding{51} & \ding{55} & \ding{55} & \ding{51} & \textbf{45.47} & 30.05 & \textbf{20.74} & \textbf{13.98} & \textbf{21.11} & \textbf{33.84} & \textbf{40.58} & \textbf{41.27} \\
\bottomrule
\end{tabular}
\end{table*}

\begin{table}[t]
\begin{minipage}[t]{0.55\linewidth} 
\centering
\footnotesize
\setlength{\tabcolsep}{0.4mm}
\begin{tabular}{lccc}
\toprule
\textbf{Method} & \textbf{Avg} & \textbf{Min} & \textbf{Max} \\
\midrule
Random              & 32.48 & 23.77 & 40.12 \\
Random (UPDP)     & 37.32 & 25.48 & 46.12 \\
Raw Report          & 39.12 & 28.16 & 46.53 \\
Raw Report (UPDP) & 41.10 & 31.44 & 47.19 \\
\bottomrule
\vspace{-2em}
\end{tabular}
\end{minipage}%
\hfill
\begin{minipage}[t]{0.42\linewidth} 
\footnotesize
\vspace*{-3\baselineskip}
\captionsetup{type=table,aboveskip=0pt}
\captionof{table}{Ablation on prompt initialization measured by the BiomedCLIP score~\cite{zhang2023biomedclip}.}
\label{tab:ablation-init}
\end{minipage}
\vspace{-1em}
\end{table}

\textbf{Evaluation for Utility-Preserving Ability During Data Transfer.} We evaluate existing mainstream models, \emph{i.e.}, Qwen2.5-VL~\cite{bai2025qwen2} on MIMIC-CXR and IU X-Ray under four training regimes: (1) using non-de-identified data only, (2) using de-identified data only, (3) using local real data only, and (4) mixed training. Table~\ref{tab:main_results1} presents a comparison between transfer data generated from raw reports, transfer data generated using our UPDP, and real data, as well as their augmentation effects on real data.

A consistent trend emerges across both datasets: de-identified transfer data generated by UPDP yields superior training performance compared to models trained on the original data, indicating that UPDP effectively preserves pathological information. For instance, under the synthetic-only training regime on MIMIC-CXR, \textbf{BLEU-1} improves from 11.34 to 13.62, while on IU X-Ray, the score rises from 42.78 to 44.42. These results demonstrate that the synthetic supervision produced by UPDP is more consistent with the paired reports and confirms its effectiveness in preserving medical semantics.

As shown in Table~\ref{tab:main_results1}, although training on transfer data alone performs slightly worse than training on local data across all evaluation metrics, the difference is small, and transfer data can also be used to augment local data. Specifically, combining local data with non-de-identified data results in better performance than training on local data alone. These results indicate that our proposed UPDP framework preserves clinically meaningful content, allowing non-de-identified data to serve as a valuable complement to local data and ultimately improve overall model performance.


Table~\ref{tab:main_results_iuxray_finetune} demonstrates that this trend is consistent across different model architectures. For Gemma-3-4B-IT, Phi-3.5-Vision-Instruct, and Qwen2.5-VL-7B-Instruct, the transfer data generated using our optimized pipeline consistently retains clinically relevant information more effectively than non-de-identified transfer data. Although the absolute performance gap relative to local data training varies among models, the consistent improvement demonstrates the utility-preserving capability of our approach across diverse backbones. Furthermore, when transfer data is combined with local data for training, the resulting performance surpasses that of local-only training, indicating that our utility-preserving transfer data can effectively complement local data.

\textbf{De-identification Effectiveness.} We evaluate the effectiveness of the proposed de-identification method by measuring the extent to which identity-related information is removed from the generated images.To this end, we construct a patient identity classification task on the \textbf{MIMIC-CXR} dataset. Specifically, we randomly select 50 patients from the dataset, resulting in a subset of 2,682 chest X-ray images. 
A classifier is trained to predict patient identity based on input images, thereby serving as a proxy for measuring identity leakage. We evaluate the trained classifier on three types of images: 
(1) local images  — classifier accuracy \textbf{97.21\%}, 
(2) images without de-identification  — classifier accuracy \textbf{13.69\%}, and 
(3) images generated by using our proposed de-identification method — classifier accuracy \textbf{4.57\%}.

If identity-related information is present in the generated images, the classifier is likely to achieve high accuracy, whereas lower accuracy reflects stronger de-identification. The results indicate that our method significantly reduces identity leakage and outperforms images produced directly from real reports. The classifier attains high accuracy on real images, confirming that patient-specific features can be learned. For images generated from real reports, accuracy remains relatively high, suggesting that some identity cues may still be preserved through report-guided generation. In contrast, accuracy drops noticeably for images generated by our method, showing that it effectively removes identity-sensitive information and prevents patient-level identification.

We further compare UPDP with representative baseline methods on IU X-Ray. As shown in Table~\ref{tab:methods_results}, UPDP attains the highest scores across all reported metrics, indicating that the generated de-identified images retain clinically relevant content and achieve a utility-preserving outcome.

\textbf{Ablation Study.} We conduct ablation analyses on three key design choices: (1) prompt initialization strategy, (2) prompt length, and (3) the number of optimization iterations. These experiments aim to identify the conditions under which content optimization yields the most substantial gains, rather than asserting exhaustive hyperparameter tuning.



\begin{table}[t]
\centering
\vspace{2mm}
\caption{Results (measure in percentage) comparison with content optimization baselines on IU X-Ray. B1 in the table stands for BLEU-1 and so forth.}
\vspace{-3mm}
\label{tab:methods_results}
\footnotesize
\setlength{\tabcolsep}{1.2mm}
\renewcommand{\arraystretch}{1.4}

\begin{tabular}{p{1.5cm}|c c c c c c}
\hline
\textbf{Method} 
& \textbf{B1} & \textbf{B2} & \textbf{B3} & \textbf{B4} 
& {\scriptsize \textbf{METEOR}} & {\scriptsize \textbf{ROUGE-L}} \\
\hline

MMA 
& 42.83 & 27.58 & 18.91 & 13.10 & 20.51 & 32.95 \\
\hline

STEPS 
& 43.43 & 28.11 & 19.19 & 13.30 & 20.06 & 33.13 \\
\hline

 UPDP
& \textbf{44.42} & \textbf{31.29} & \textbf{19.90} & \textbf{13.84} & \textbf{21.08} & \textbf{33.58} \\
\hline

\end{tabular}
\vspace{-4mm}
\end{table}

\begin{figure*}
\centering
\includegraphics[width=0.9\linewidth]{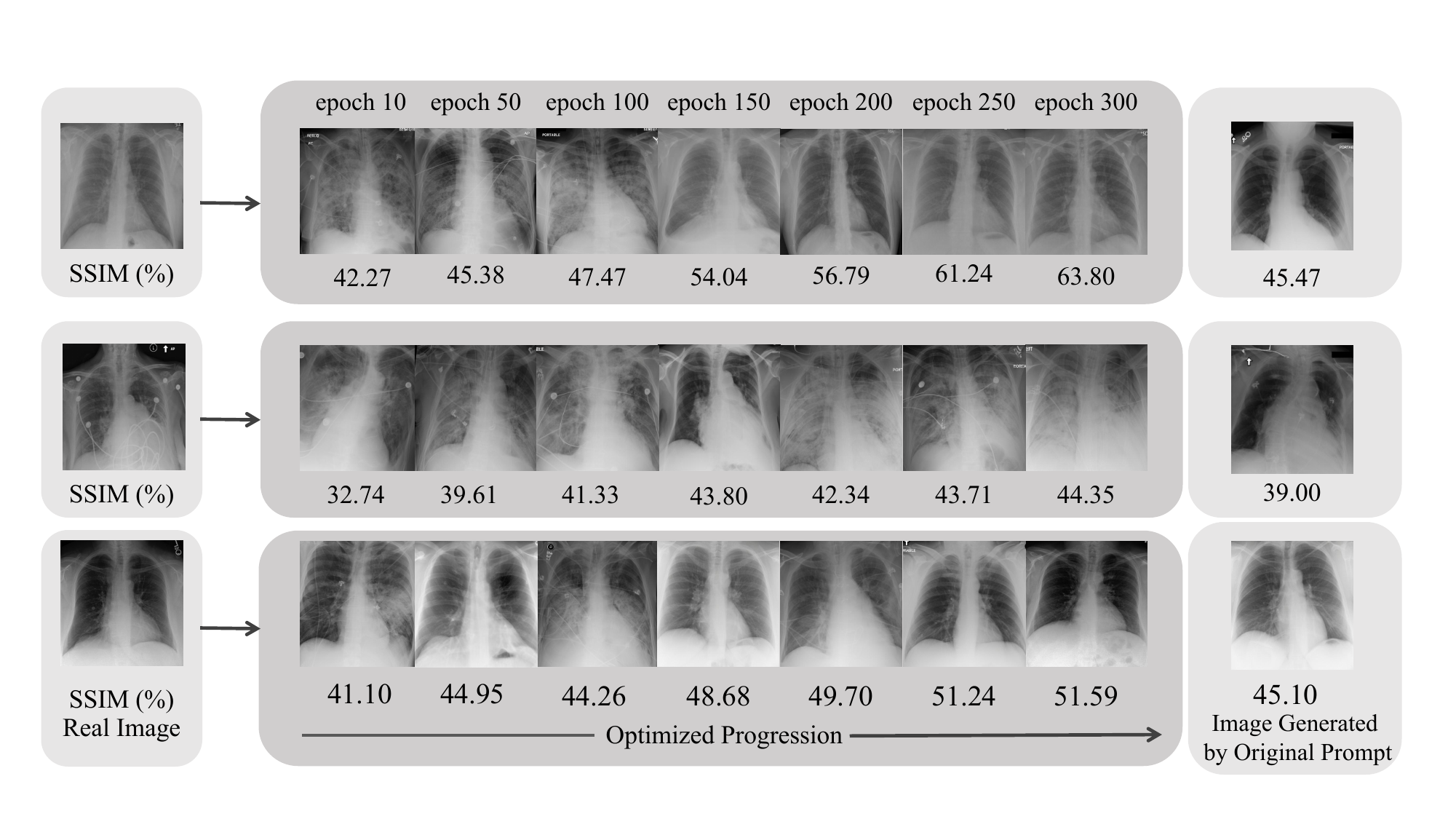}
\vspace{-2em}
\caption{Comparison of generated chest X-rays across different content optimization iterations with corresponding SSIM (\%) values against real images.}
\vspace{-1.0em}
\label{fig:intro}
\end{figure*}

\begin{figure}[t]
\centering
\begin{minipage}[c]{0.28\textwidth}
\centering
\includegraphics[width=\textwidth]{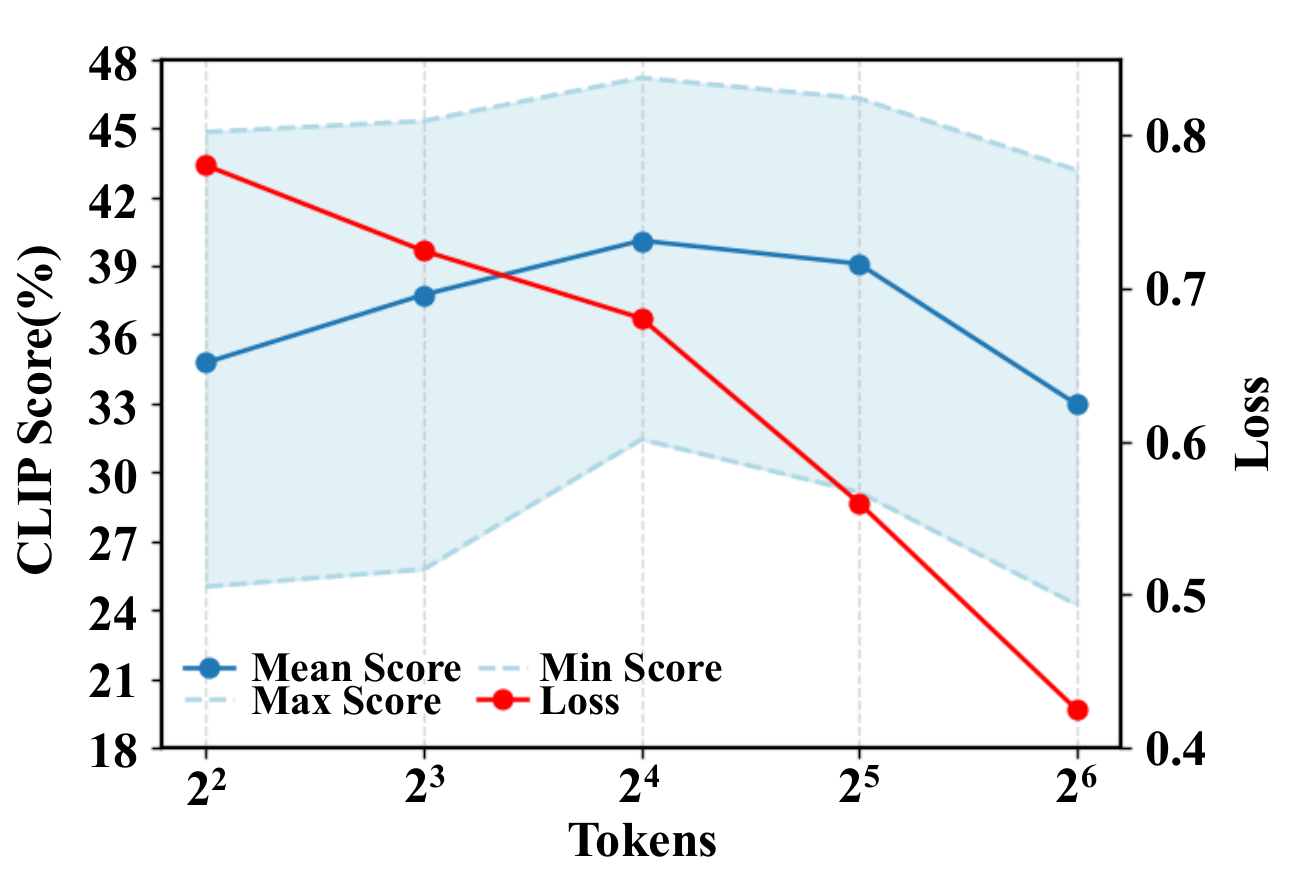}
\end{minipage}%
\begin{minipage}[c]{0.20\textwidth}
\vspace*{-1\baselineskip}
\caption{Effect of prompt content length on training loss and the CLIP score. The CLIP score follows an inverted U-shaped curve with increasing token count, while training loss continues to drop.}
\label{fig:token}
\end{minipage}
\vspace{-2em}
\end{figure}

\textbf{Effect of token length and BiomedCLIP evaluation.}
We analyze the effect of prompt length by varying the number of tokens in the optimized content and measuring the BiomedCLIP score~\cite{zhang2023biomedclip} between the de-identified images and corresponding reports. Figure~\ref{fig:token} reveals an inverted U-shaped relationship: scores improve up to a moderate token count before declining, even as training loss continues to drop. This divergence suggests that moderate prompt lengths achieve a more favorable trade-off between representational capacity and generalization, whereas excessively long prompts may introduce redundancy or overfitting that harms cross-modal alignment.


\textbf{Effectiveness of Content Optimization and Iteration.}
We study how iterative content optimization (\textit{i.e.}, multiple iterations of content optimization) affects the quality of generated, de-identified chest X-rays. As shown in Figure~\ref{fig:intro}, the refined instructions progressively guide the generation process through multiple content optimization iterations, with image fidelity improving noticeably as the number of iterations increases, suggesting a cumulative effect of iterative refinement. Quantitative analysis via SSIM scores confirms this trend, showing steady enhancement in structural similarity to real images across iterations, which indicates that the iterative process effectively preserves clinically relevant features. Importantly, even when improvements plateau at higher numbers of content optimization iterations, the refined instructions produce superior image quality compared to initial or unoptimized versions, demonstrating the lasting impact of careful content design. These results highlight that both the formulation of instructions and iterative optimization play a critical role in generating realistic, diagnostically meaningful, and fully de-identified chest X-rays.

\begin{figure*}[t]
\centering
\includegraphics[width=1\textwidth]{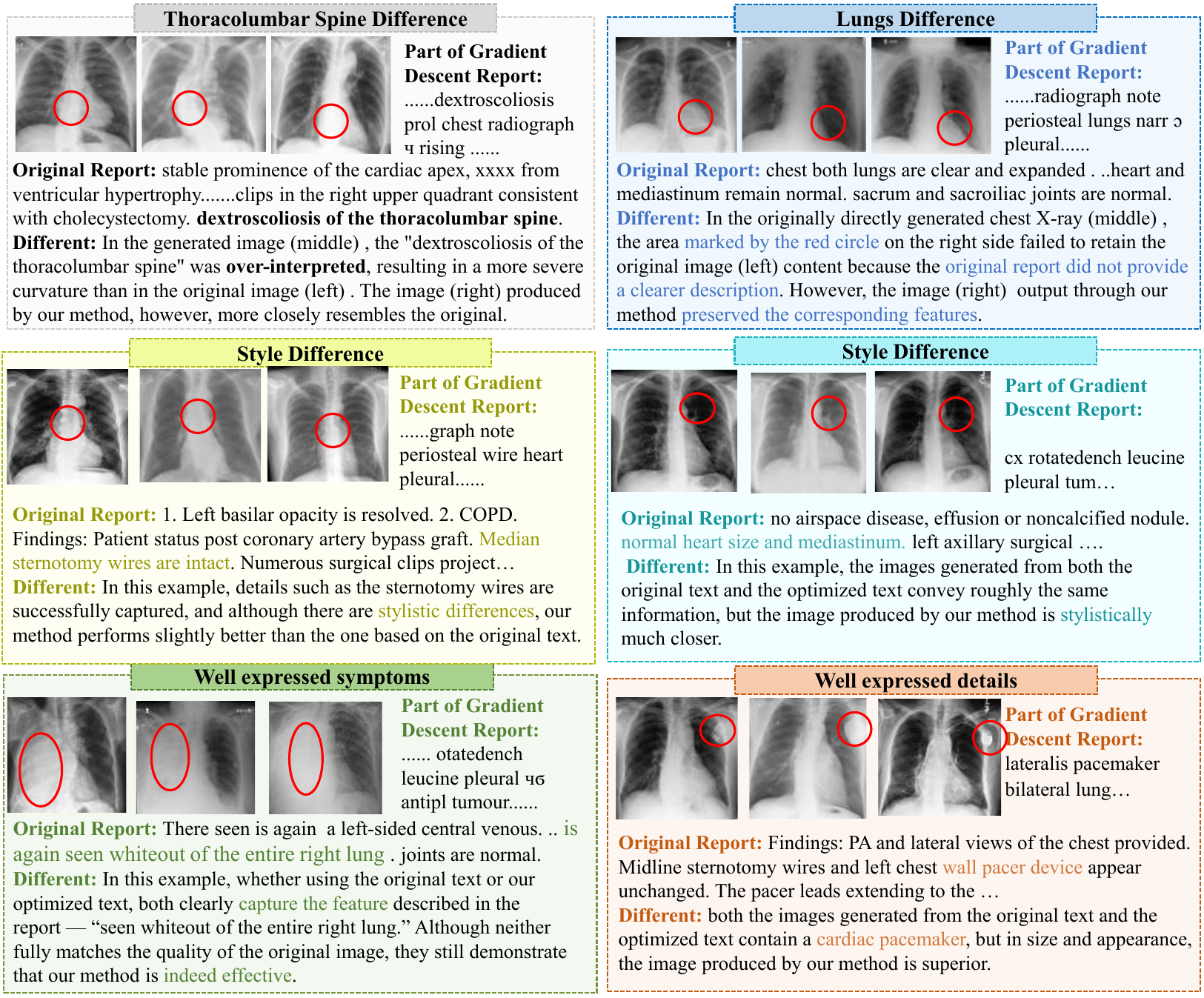}
\vspace{-8mm}
\caption{
Qualitative comparison of synthetic chest X-rays generated from raw reports and optimized contents. Optimized contents better preserve anatomy, radiographic style, and several fine-grained findings in the examples shown here.}
\label{fig:qualitative_results}
\vspace{-1em}
\end{figure*}
\textbf{Effect of prompt initialization.} We compare four initialization strategies: random initialization, optimized contents derived from random initialization, raw reports, and optimized contents initialized from raw reports. Table~\ref{tab:ablation-init} shows that content optimization provides benefits in both initialization regimes, improving overall generation quality, but the best results are achieved when optimizing prompts that start from the raw report. This finding suggests that semantic initialization not only simplifies the subsequent optimization search but also facilitates stronger alignment between the generated images and their corresponding textual descriptions.

Overall, these ablation results suggest that the effectiveness of content optimization is not intrinsic, but rather contingent on a set of practical design choices, including prompt length, initialization strategy, and the number of optimization iterations. Variations along these dimensions lead to non-trivial differences in downstream performance, indicating that content optimization must be carefully configured and tuned, rather than being treated as a simple plug-and-play component in the generation pipeline.

Within the reported setting, a clear pattern emerges: the method achieves its strongest performance when the prompt is initialized from the original report and constrained to a moderate token length. This configuration likely provides a balanced trade-off—retaining sufficient semantic grounding from the source report while avoiding redundancy or noise introduced by overly long prompts.

\vspace{3em}

\subsection{Qualitative Results}
Figure~\ref{fig:qualitative_results} presents qualitative comparisons of generated images under different training settings. 
Images generated from non-de-identified reports may omit certain anatomical structures or exhibit stylistic deviations from real X-rays, such as variations in tissue contrast, organ boundaries, or the representation of subtle pathological signs. In contrast, images generated from optimized contents consistently better preserve organ structures, radiographic appearance, and multiple subtle findings that are clinically relevant, including fine-grained details that are important for diagnosis. These qualitative observations illustrate that our approach not only improves visual fidelity but also maintains critical diagnostic information, supporting the utility-preserving objective of the synthetic data generation.

\vspace{-3mm}
\subsection{Discussion}

Overall, our results indicate that de-identified transfer data generated by UPDP yields superior training outcomes for report generation compared to synthetic data derived from non-de-identified sources. However, since a slight performance gap remains compared to local real-data training, we recommend treating de-identified data as a privacy-preserving augmentation strategy, or as a fallback option under data-access constraints.

Nevertheless, de-identified data remains advantageous due to its inherent scalability: once the de-identification pipeline is established, large-scale samples can be generated efficiently. This scalability not only facilitates systematic augmentation of training datasets, but also enables rebalancing of training distributions and supports experimentation in low-resource scenarios.

Moreover, mixed training that combines optimized de-identified data with a small amount of real data consistently achieves the best performance in our experiments. This suggests that the two data sources play complementary roles during training. Specifically, de-identified data can substantially increase the diversity and coverage of training samples, while even a relatively small amount of real data provides reliable visual grounding and prevents the model from over-relying on features introduced during de-identification~\cite{kebaili2025multi,frid2018gan}. As a result, this hybrid strategy offers a practical compromise between scalability and realism when obtaining large-scale annotated medical images is challenging.



This study has several limitations. Although traditional text-overlap metrics and semantic or structured measures such as BERT\allowbreak Score and RadGraph were employed, these automated evaluations mainly capture surface-level similarities and lack expert assessment. Consequently, they are insufficient to fully evaluate clinical utility, decision-making relevance, or potential privacy risks. Future work should establish a comprehensive validation framework that integrates stronger medical priors in content optimization, incorporates expert evaluation to assess diagnostic accuracy and usability, and conducts explicit privacy analyses to quantify potential information leakage or membership inference risks.

\section{Conclusion}

We propose a utility-preserving de-identification pipeline (UPDP) for secure cross-hospital sharing of radiology data. By employing a privacy-sensitive term blacklist, a pathology-related whitelist, and generative filtering, UPDP generates privacy-filtered, pathology-preserving synthetic images and de-identified reports, thereby maintaining diagnostic value while safeguarding patient identity. Experiments on public chest X-ray datasets demonstrate that UPDP effectively removes identity information while preserving critical pathological features. Models trained on de-identified data achieve diagnostic performance comparable to models trained on original data, with reduced identity-related accuracy, confirming effective privacy protection. In cross-hospital scenarios, integrating de-identified data with local data further improves performance, validating the safety and utility of UPDP. Overall, UPDP offers a practical solution for privacy-compliant sharing of large-scale radiology data, supporting cross-institutional medical AI development and evaluation.

\vspace{-1mm}
\begin{acks}
This work was supported in part by the Shandong Province Overseas Young Talents Program (2026HWYQ-009) and the Key Research and Development Program of Shandong Province  (2025CXGC010901).
\end{acks}

\bibliographystyle{ACM-Reference-Format}
\bibliography{main}

\appendix









\end{document}